\documentclass[a4paper,twocolumn]{article}

\usepackage[utf8]{inputenc}
\usepackage{eurosis}
\usepackage{graphicx}
\usepackage{myHarvard}
\usepackage{amsmath}
\usepackage{algorithmic}

\title{LEARNING A REPRESENTATION OF A BELIEVABLE VIRTUAL CHARACTER'S ENVIRONMENT WITH AN IMITATION ALGORITHM}
\author{Fabien Tencé$^{*,**}$, Cédric Buche$^*$, Pierre De Loor$^*$ and Olivier Marc$^{**}$\\
$^*$ UEB -- ENIB -- LISyC\\
$^{**}$ Virtualys\\
Brest -- France\\
\{tence,buche,deloor\}@enib.fr, olivier.marc@virtualys.com}
\date{}

\begin{document}

\maketitle
\thispagestyle{empty}


\begin{abstract} 
In video games, virtual characters' decision systems often use a simplified representation of the world. To increase both their autonomy and believability we want those characters to be able to learn this representation from human players. We propose to use a model called \emph{growing neural gas} to learn by imitation the topology of the environment. The implementation of the model, the modifications and the parameters we used are detailed. Then, the quality of the learned representations and their evolution during the learning are studied using different measures. Improvements for the growing neural gas to give more information to the character's model are given in the conclusion. 
\end{abstract}

\section{INTRODUCTION}
\paragraph{}
One of the major goals of video games is to make the user feel like he/she really is in the game environment. To achieve this, it is possible to use complex devices to increase \emph{immersion} like surround sound systems and stereoscopic screens. Another way is to make rich environments to increase \emph{presence}. The latter can be done by, among other things, having characters with believable behaviours \cite{Bates1994}. To generate those behaviours, models from artificial intelligence are used in the game industry. For those models to perceive the environment, game designers have to create a simplified representation of the environment, which is often defined \emph{a priori}.
\paragraph{}
Defining by hand every new environment's representation is a time-consuming work. We propose that our characters will be able to learn those representations, for them to be autonomous. This learning will be unsupervised and online: the character will learn while it plays without the judgement of a human. To be able to achieve the best believability, we want the computer-controlled characters to do like human-controlled characters. Indeed, there are no better example of what a believable behaviour is than a human behaviour itself. It is this kind of learning, by example \cite{Del_Bimbo_95} or by imitation \cite{Gorman2007,Bauckhage2007} we want to use to learn a representation of the environment.
\paragraph{} 
This article first presents the \emph{growing neural gas} model which is used to learn the representation of the environment. Then the characteristics and qualities of the learned representations are assessed by different measures. To conclude, some enhancements are proposed for the growing neural gas to give more information to the character's model.

\section{LEARNING A REPRESENTATION OF THE ENVIRONMENT}
\paragraph{} 
Models which control virtual characters use different types of representation to find paths to go from one point to another. Classic approaches use a graph: nodes represent accessible places and edges represent paths between each place. Actual solutions tend to use a mesh, with different degrees of complexity, to represent the zones where the character can go. The problem with the latter solutions is that they use algorithms to find the optimal path between two points and not the most believable path. With graphs, it is possible to have more control on the paths used by the character.
\paragraph{} 
Instead of designing graphs \textit{a priori}, we want them to be learned by imitation of a human player. This work as been done in \cite{Thurau2004a} where nodes and a potential field are learned from humans playing a video game. The character is then using this representation to move in the game environment, following the field defined at each node. To learn the position of the nodes, Thurau use an algorithm called Growing Neural Gas (GNG).
\paragraph{} 
The GNG \cite{Fritzke1995} is a graph model which is capable of incremental learning. Each node has a position (x,y,z) in the environment and has a cumulated error which measures how well the node represents its surroundings. Each edge links two nodes and has an age which gives the time since it was last activated. This algorithm needs to be omniscient, because the position of the imitated player, the demonstrator, is to be known at any time.
\paragraph{} 
The principle of the GNG is to modify its graph, adding or removing nodes and edges and changing the nodes' position for each given demonstrator's position. The algorithm does the following: for each input the closest and the second closest nodes are picked. An edge is created between those nodes and the closest node's error is increased. Then the closest node and its neighbours are attracted toward the input. The ages of the closest node's edges are increased by 1 and too old edges are deleted. Each $\lambda$ input a node is inserted between the node with the maximum error and its neighbours having the maximum error. At the end of an iteration, each node's error is decreased by a small amount.
\paragraph{} 
The version we use is a bit modified to give better results for our needs as shown by figure \ref{algoGNG}. Instead of inserting a new node each $\lambda$ input, a node is inserted when a node's error is superior to a parameter $MAX\_ERROR$. As each node's error is reduced by a small amount $ERROR\_DECAY$ for each input, the modified GNG algorithm does not need a stopping criterion. Indeed, if there are many nodes which represent well the environment, the error added for the input will be small and for a set of inputs, the total added error will be distributed among several nodes. The decay of the error will avoid new nodes to be added to the GNG resulting in a stable state. However if the player which serves as a example goes to a place in the environment he has never gone before, the added error will be enough to counter the decay of the error, resulting in new nodes to be created.
\begin{figure}[!ht]
\begin{tabular}{|c|}
\hline
\begin{minipage}{\linewidth}
\vspace{0.2cm}
\begin{algorithmic}
\STATE nodes $\gets$ \{\}
\STATE edges $\gets$ \{\}
\WHILE {demonstrator plays}
  \STATE (x,y,z) $\gets$ demonstrator's position
  \IF {$|$nodes$|$ = 0 or 1}
    \STATE nodes $\gets$ nodes $\cup$ \{(x,y,z,error=0)\}
  \ENDIF
  \IF {$|$nodes$|$ = 2}
    \STATE edges $\gets$ \{(nodes,age=0)\}
  \ENDIF
  \STATE first $\gets$ closest((x,y,z),nodes)
  \STATE second $\gets$ secondClosest((x,y,z),nodes)
  \STATE edge $\gets$ edges $\cup$ \{\{first,second\},age=0)\}
  \STATE
  \STATE first.error+=$||$(x,y,z)-first$||$
  \STATE Attract first toward (x,y,z)
  \STATE $\forall$ edge $\in$ first's edges, edge.age++
  \STATE Delete edges older than MAX\_AGE
  \STATE Attract neighbours(first) toward (x,y,z)
  \STATE $\forall$ node $\in$ nodes, node.error-=ERROR\_DECAY
  \STATE
  \IF {first.error $>$ MAX\_ERROR}
    \STATE maxErrNei $\gets$ maxErrorNeighbour(first)
    \STATE newNode $\gets$ between(first,maxErrNei)
    \STATE first.error/=2, maxErrNei.error/=2
    \STATE newError $\gets$ first.error+maxErrNei.error
    \STATE nodes $\gets$ nodes $\cup$ \{(newNode,newError)\}
  \ENDIF
\ENDWHILE
\end{algorithmic}
\vspace{0.2cm}
\end{minipage}\\
\hline
\end{tabular}
\caption{Algorithm used to learn the topology of the environment represented by a growing neural gas.}
\label{algoGNG}
\end{figure}
\paragraph{} 
This algorithm has 5 parameters which influence the density of nodes, the quality of the representation, the adaptivity and the time to reach a stable state:
\begin{itemize}
 \item The attraction applied to $first$ toward $(x,y,z)$
 \item The attraction applied to $first$'s neighbours toward $(x,y,z)$
 \item The nodes' error decay, $ERROR\_DECAY$
 \item The nodes' maximum error, $MAX\_ERROR$
 \item The edges' maximum age, $MAX\_AGE$
\end{itemize}
\paragraph{} 
Nodes can be used as a representation for the character's decision system. However, edges only represent proximity and not paths between nodes: nodes can be close but there may be a obstacle between them, so edges cannot be used by the model.

\section{EVALUATION}
\begin{figure}[!ht]
  \centering
  \includegraphics[width=0.69\linewidth]{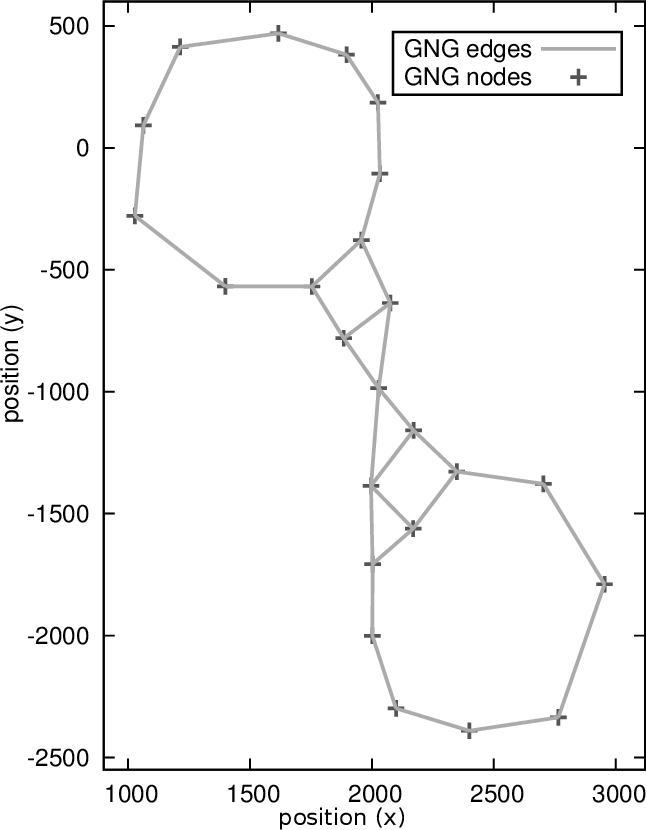} 
  \caption{Result of a growing neural gas learned from a player for a simple map, top view.}
  \label{2D_fullGNG}
\end{figure}

\begin{figure}[!ht]
  \centering
  \includegraphics[width=0.85\linewidth]{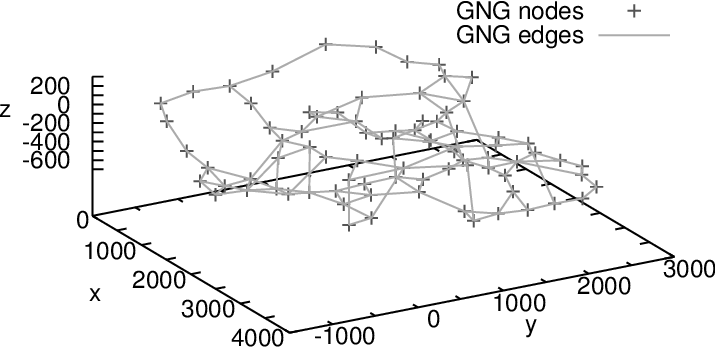} 
  \caption{Result of a growing neural gas learned from a player for a complex map.}
  \label{3D_fullGNG}
\end{figure}

\begin{figure}[!ht]
  \centering
  \includegraphics[width=0.68\linewidth]{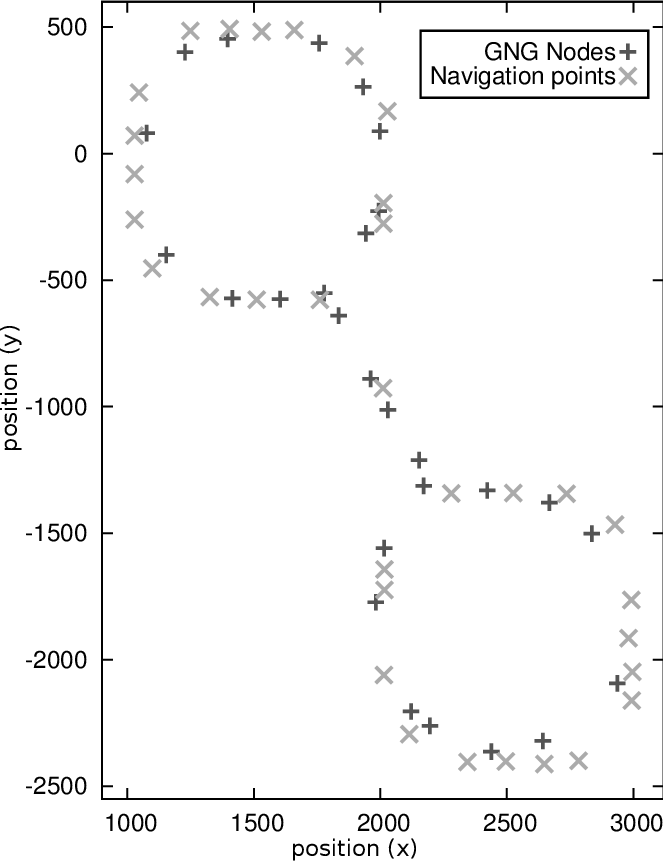} 
  \caption{Comparison of nodes learned by the growing neural gas with the navigation points placed manually by the game developers.}
  \label{comp_navs_gng}
\end{figure}

\begin{figure}[!ht]
  \centering
  \includegraphics[width=0.92\linewidth]{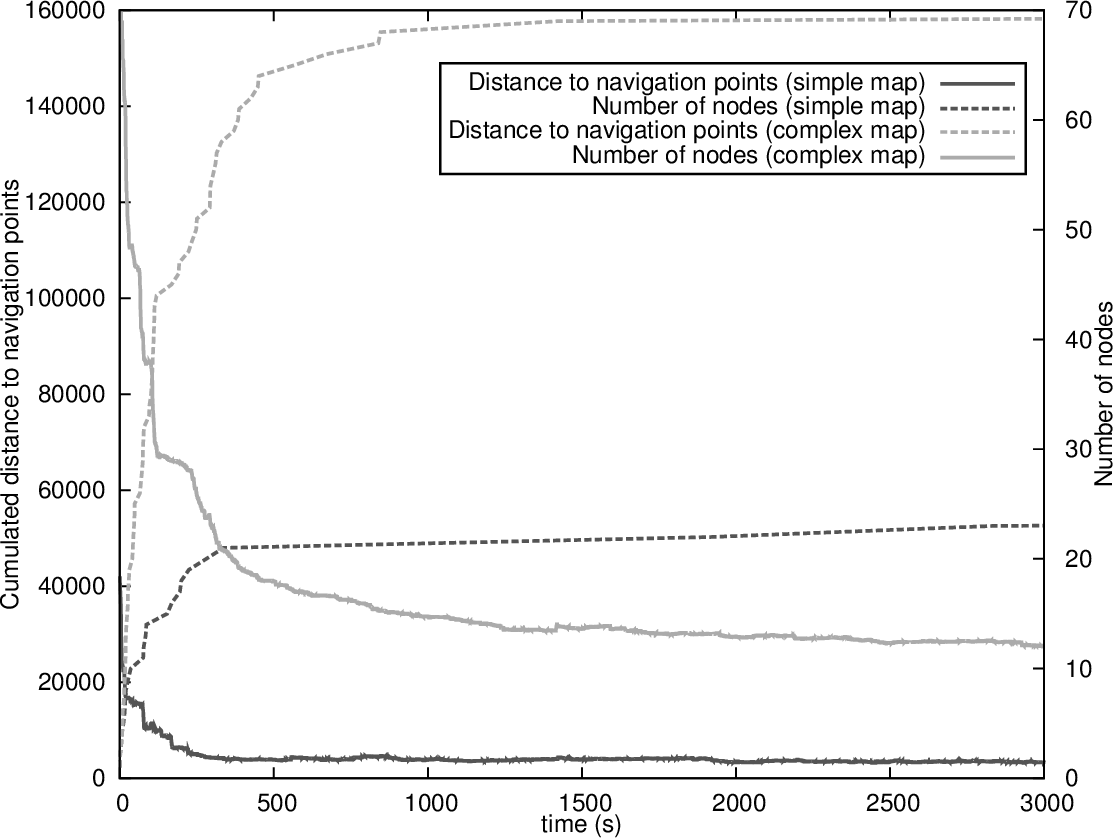} 
  \caption{Time evolution of the cumulated distance to navigation points defined manually and the growing neural gas' nodes and the growing neural gas' number of nodes.}
  \label{convergenceTime}
\end{figure}

\begin{figure}[!ht]
  \centering
  \includegraphics[width=0.92\linewidth]{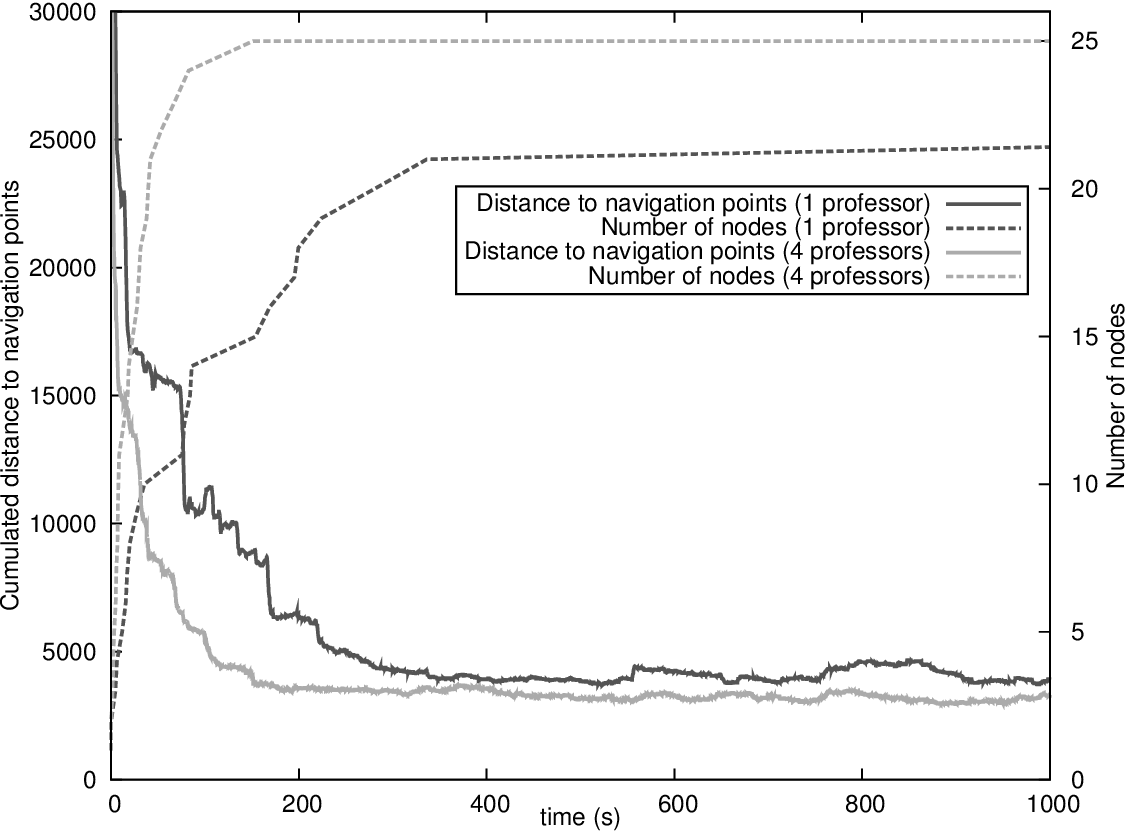} 
  \caption{Time evolution of the cumulated distance to navigation points defined manually and the growing neural gas' nodes and the growing neural gas' number of nodes.}
  \label{convergenceMultiProf}
\end{figure}

\begin{figure}[!ht]
  \centering
  \includegraphics[width=0.68\linewidth]{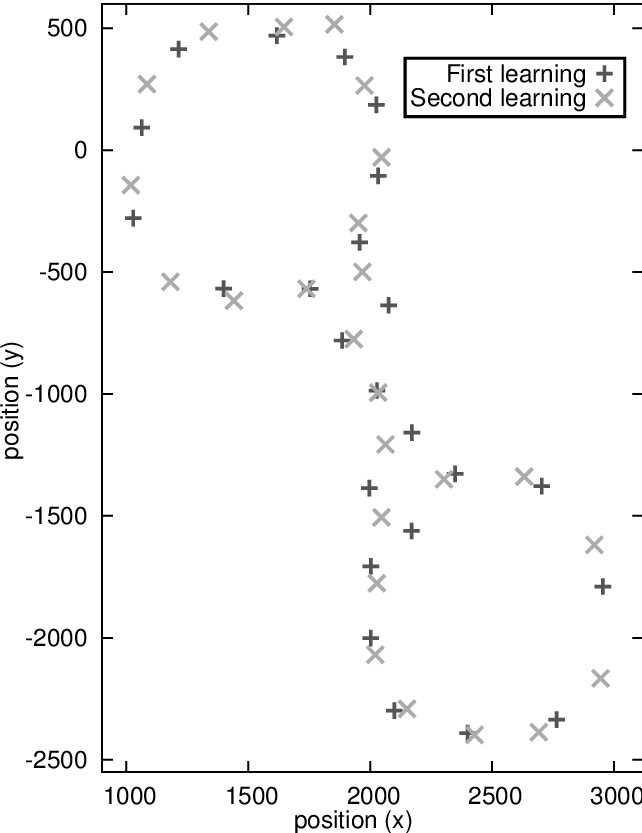} 
  \caption{Comparison of two growing neural gas which learned on the same environment, after a very long training time (more than 10 hours).}
  \label{convergenceNodes}
\end{figure}

\paragraph{} 
We used the game Unreal Tournament 2004 because it features quite complex environments. Human players can also control avatars in the game so the GNG can learn for them. We have to choose the parameters in a empirical way because we cannot find them analytically nor use an optimization algorithm. Indeed, our goal is believability and it can be only measured with human judges. This kind of evaluation is not suitable for optimization. The best parameters we found are:
\begin{itemize}
 \item Attraction force applied to $first$ is 0.03 times the vector $(x,y,z)-first$
 \item Attraction force applied to $first$'s neighbours is 0.0006 times the vector $(x,y,z)-second$
 \item Nodes' error decay is 10
 \item Nodes' maximum error is 20000
 \item Edges' maximum age is 75
\end{itemize}
To compare with other environments, the position in Unreal Tournament is given in Unreal units (1 meter is roughly equal to 50 Unreal units) and all the parameters are based on a demonstrator's position in Unreal units.
\paragraph{} 
With those parameters we trained 2 GNG on 2 different maps. The first one is a simple map, called Training Day. It is small and flat which is interesting to visualize the data in 2 dimensions. The second one, called Mixer, is much bigger and complex with stairs, elevators and slopes which is interesting to see if the GNG behave well in 3 dimensions. The results is given in figure \ref{2D_fullGNG} for the simple map and in figure \ref{3D_fullGNG} for the complex map.
\paragraph{} 
To study the quality of the learned topology, we first choose to compare the GNG's nodes with the navigation point placed manually by the map creators. Of course, we do not want the GNG to fit exactly those points but it gives a first evaluation of the learned representation. In our case, we have those navigation points but our goal is that they are not longer necessary for a character to move in a new environment. Figure \ref{comp_navs_gng} shows both the navigation points and the GNG's nodes. As we can see, the two representations look alike which indicates that the model is very effective in learning the shape of the map. However, there are zones where the GNG's nodes are more concentrated than the navigation points and other where they are less concentrated. We cannot tell now if it is a good behaviour or not as we should evaluate an agent using this representation to see if it navigate well. Even in the less concentrated zones, the nodes are always close enough to be seen from one to another, so it should not be a problem.
\paragraph{} 
As the attraction applied to the nodes for each input is constant, the GNG is not converging to a fixed state. This is a wanted behaviour, allowing the GNG to adapt to a variation in the use of the map: if the demonstrator suddenly uses a part of the map which he/she has not explored yet, the GNG will be able to learn this new part even if the GNG has been learning for a long time. We do want, however, the GNG to learn quickly the topology and to keep a good representation of the world over time.
\paragraph{} 
To study the time evolution of the GNG's characteristics, we introduce a distance measure: the sum of the distance between each navigation point and its closest node. We also study the evolution of the number of nodes because we do not want the GNG to grow indefinitely. Figure \ref{convergenceTime} shows this two measures for the simple and the complex maps. For the simple map, the GNG reached its maximum number of node and minimum error in approximatively 5 minutes of real-time simulation. For the complex map, it takes more time, about 25 minutes, but results at 12 minutes are quite good. Those results show that it is possible to have an character that learns during the play.
\paragraph{} 
The GNG can handle inputs from multiple demonstrators. Figure \ref{convergenceMultiProf} shows the distance and number of node for a GNG trained on 1 demonstrator and for a GNG trained on 4 demonstrators. The learning with 4 demonstrators is, as expected, faster: about 3 minutes for the distance to stabilize instead of 5 minutes for 1 demonstrator. It is interesting to note that the learning is not 4 times faster but the gain is still important. Learning with multiple demonstrators seems to give a GNG with less variation during the learning. The gain have however a small drawback: the number of nodes is a little superior for multiple demonstrators. It may be due to the fact that demonstrators are scattered in the environment instead of a unique demonstrator following a path.
\paragraph{} 
It is interesting to compare two learned GNG on the same demonstrator in the same environment and conditions but for 2 different simulations. The goal is to see if the two representations fit. Figure \ref{convergenceNodes} shows that the resulting GNG are a bit different. The first GNG has 24 nodes and has a cumulated distance to navigation points of approximatively 3300 Unreal units. The second GNG has 25 nodes and has a cumulated distance of approximatively 3150 Unreal units. This proves that the GNG does not converge toward a unique solution but those solutions are quite similar in shape, number of nodes and distance to navigation points. 
\paragraph{} 
The last evaluation assesses the impact of the frequency at which the demonstrator's position is given to the GNG. For the previous experiments, the frequency was set to 10Hz. Figure \ref{comp_freq} shows the differences for 1, 10 and 100 Hz. Results indicate that 1Hz give comparable results to 10Hz but it takes much longer to give a good representation. At 100Hz, the GNG reaches a stable state as fast as at 10Hz but the resulting GNG has much more nodes resulting in a lower error. Using a high frequency is therefore not useful because the number of nodes can be increased changing the $MAX\_ERROR$ and $ERROR\_DECAY$ parameters without using computing power.
\begin{figure}[!ht]
\begin{tabular}{|l|l|l|l|}
\hline
Frequency&Time&Number of nodes&Error\\
\hline
1Hz&1h30&22&3800 UU\\
10Hz&5min&24&3300 UU\\
100Hz&5min&39&2300 UU\\
\hline
\end{tabular}
\caption{Comparison of growing neural gas' characteristics learned at different frequencies on a simple map (Training Day). Time, number or nodes and error are given when the growing neural gas reach a stable state. UU stands for Unreal units.}
\label{comp_freq}
\end{figure}

\section{IMPROVEMENTS}
\paragraph{} 
The main drawback of the GNG is that the only information extracted is the position of the nodes. The biggest difference between the GNG we implemented and navigation graphs coded manually is that the latter give also information on the accessibility of a node from another. Edges in a GNG gives only an information on proximity but there can be a obstacle between two nodes joined by an edge. An idea could be to store the previously activated node and create an edge to the current activated node. Like the GNG edges we should make the edge age and disappear if they are too old. Whether those edges should replace the GNG edges could be an interesting experiment to set up.
\paragraph{} 
To share more information with the behaviour model, we can learn which kind of action is done when the character is near a node. The character, knowing its nearest node, can choose the best action to do. This process is quite similar to the process of tagging: designers often annotate navigation points with information such as ``jump spot'' or ``covering''.
\paragraph{} 
Another interesting information to learn is if the demonstrator walk exactly at the node's position or if there is a big variation in the distance to the node. The nodes' error give a bit of information about this variation, however with the error decay, this information is lost over time. The learning is problematic because each winner node moves so it does not represent the same area. It could be possible to update the influence radius of winner and its neighbours according to their current radius and the distance to the example. The exact formula is yet to be found.

\section{CONCLUSION}
\paragraph{}
Virtual environments, like for example video games, need believable characters for users to feel in the environment. To improve the characters' behaviour, we decided to use a growing neural gas to learn by imitation the topology of the environment. We believe that it will make the agent use the environment in a more human-like fashion. It also removes the burden from the maps designers of placing manually the navigation graph.
\paragraph{}
Our first evaluations tend to show that the growing neural gas gives a good representation of the environment. The learning is fast, with one demonstrator it takes up to 25 minutes to learn a representation of the whole environment. As it is possible to learn with several demonstrators, learning can be done very quickly. The character can thus adapt quickly to changes in the use of the environment and evolve while playing. Although different runs gives different results, the representations are very similar. 
\paragraph{}
With this ability to learn the environment, the agent can be placed in any simulation without \textit{a priori} knowledge and still be able to move by imitating human users. As the learning is quite fast, users could perceive the evolution in the way the agent acts and thus believing it can be human. The growing neural gas gives autonomy and believability to the model.
\paragraph{}
The next step is to put more information in the growing neural gas, learning which node is accessible from each node and finding the best action at each node. To see if this work really gives results we will have to test the difference in the behaviour between an agent using the navigation points and an agent using the growing neural gas.

\bibliographystyle{myBibStyle}
\bibliography{bibCB,library}

\end{document}